\documentclass[letterpaper, 10 pt, journal, twoside]{IEEEtran} 
\usepackage{graphicx,psfrag,amsmath,amsfonts,verbatim,url,booktabs,array,float}
\usepackage[small,bf]{caption}

\usepackage{algorithm}
\usepackage{algpseudocode}
\usepackage{url}

\usepackage[top=60pt,bottom=60pt,rmargin=50pt,lmargin=50pt]{geometry}

\usepackage{pgfplots}
\pgfplotsset{compat=1.15,
	legend style={font=\footnotesize},
}
\usepackage{tikzscale}
\usetikzlibrary{matrix}
\usetikzlibrary{backgrounds}
\usepgfplotslibrary{groupplots}

\pgfplotsset{every axis/.append style={
		scaled y ticks = false,
		scaled x ticks = false,
		y tick label style={/pgf/number format/.cd, fixed, fixed zerofill,
			int detect,1000 sep={\;},precision=2},
	},
	legend image code/.code={
		\draw[mark repeat=2,mark phase=2]
		plot coordinates {
			(0cm,0cm)
			(0.1cm,0cm)        
			(0.1cm,0cm)         
		};%
	}
}

\makeatletter
\newcommand{\linebreakand}{%
\end{@IEEEauthorhalign}
\hfill\mbox{}\par
\mbox{}\hfill\begin{@IEEEauthorhalign}
}
\makeatother

\begin{document}
	
	\title{Trajectory Optimization \\ with Optimization-Based Dynamics}
	
	\author{Taylor A. Howell$^{1}$, Simon Le Cleac'h$^{1}$, Sumeet Singh$^{2}$, Pete Florence$^{2}$, Zachary Manchester$^{3}$, Vikas Sindhwani$^{2}$
		\thanks{$^{1}$Taylor A. Howell and Simon Le Cleac'h are with the Department of Mechanical Engineering,
			Stanford University,
			Stanford, CA 94305, USA
			{\tt\small \{thowell,simonlc\}@stanford.edu}}%
		\thanks{$^{2}$ Sumeet Singh, Pete Florence, and Vikas Sindhwani are with Robotics at Google,
			New York City, NY 10011 and Mountain View, California 94043 USA
			{\tt\small \{ssumeet,peteflorence,sindhwani\}@google.com}}%
		\thanks{$^{3}$Zachary Manchester is with the The Robotics Institute,  Carnegie Mellon University,
			Pittsburgh, PA 15213 USA
			{\tt\small zacm@cmu.edu}}%
		\thanks{This work was supported by Google Research.}
	}

	\maketitle
	\begin{abstract}
		We present a framework for bi-level trajectory optimization in which a system's dynamics are encoded as the solution to a constrained optimization problem and smooth gradients of this lower-level problem are passed to an upper-level trajectory optimizer. This optimization-based dynamics representation enables constraint handling, additional variables, and non-smooth behavior to be abstracted away from the upper-level optimizer, and allows classical unconstrained optimizers to synthesize trajectories for more complex systems. We provide an interior-point method for efficient evaluation of constrained dynamics and utilize implicit differentiation to compute smooth gradients of this representation. We demonstrate the framework by modeling systems from locomotion, aerospace, and manipulation domains including: acrobot with joint limits, cart-pole subject to Coulomb friction, Raibert hopper, rocket landing with thrust limits, and planar-push task with optimization-based dynamics and then optimize trajectories using iterative LQR.
	\end{abstract}

	
	\section{Introduction}
	\IEEEPARstart{T}{rajectory} optimization is a powerful tool for synthesizing trajectories for nonlinear dynamical systems. Indirect methods, in particular, are able to efficiently find optimal solutions to this class of problem by returning dynamically feasible trajectories via rollouts and utilizing gradients of the system's dynamics.
	
	Classically, indirect methods like iterative LQR (iLQR) \cite{jacobson1970differential} utilize \textit{explicit} dynamics representations, which can be directly evaluated and differentiated in order to return gradients. In this work, we present a more general framework for \textit{optimization-based} dynamics representations. This latter class enables partial elimination of trajectory-level constraints by absorbing them into the dynamics representation.

	\begin{figure}[t]
		\centering
		\includegraphics[width=.3\textwidth]{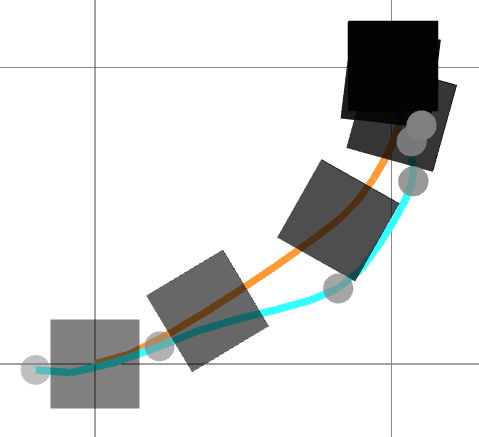}
		\caption{Optimized trajectories for planar-push task. The pusher (blue) and block (orange) paths are shown for a plan that maneuvers the block from a pose at the origin to a goal pose.}
		\label{planar_push}
	\end{figure}

	We formulate optimization-based dynamics as a constrained optimization problem and provide an interior-point method for efficient evaluation of the dynamics at each time step. Implicit differentiation is utilized to compute derivatives for this representation, and we exploit intermediate results from our interior-point method to return useful, smooth gradients to an upper-level optimizer.
	
	To demonstrate the capabilities of this representation, we utilize optimization-based dynamics and iLQR in a bi-level optimization framework for planning. We provide a number of optimization-based dynamics models and examples in simulation including: an acrobot with joint limits, a cart-pole experiencing friction, gait generation for a Raibert hopper, a belly-flop soft landing for a rocket subject to thrust limits, and a planar-push manipulation task. We compare our approach to MuJoCo, contact-implicit trajectory optimization, and gradients generated with randomized smoothing.
	
	Specifically, our contributions are:
	\begin{itemize}
		\item A novel framework for optimization-based dynamics that can be used with trajectory optimization methods that require gradients
		\item An interior-point method that can efficiently evaluate constrained optimization problems and return smooth gradients
		\item A collection of optimization-based dynamics models and examples from locomotion, aerospace, and manipulation domains that demonstrate the proposed bi-level trajectory optimization framework
	\end{itemize}
	
	In the remainder of this paper, we first review related work on bi-level approaches to trajectory optimization in Section \ref{related_work}. Next, we present background on the iLQR algorithm, implicit differentiation, and implicit integrators in Section \ref{background}. Then, we present our optimization-based dynamics representation, as well as an interior-point method for solving this problem class in Section \ref{id}. We provide a collection of optimization-based dynamics models that are utilized to perform trajectory optimization, and comparisons, in Section \ref{examples}. Finally, we conclude with discussion of limitations and directions for future work in Section \ref{conclusion}.
	
	\section{Related Work}\label{related_work}
	Bi-level optimization \cite{sinha2017review} is a framework where an upper-level optimizer utilizes the results, i.e., solution and potentially gradients, of a lower-level optimization problem. Approaches typically either implicitly solve the lower-level problem and compute gradients using the solution, or explicitly represent the optimality conditions of the lower-level problem as constraints in the upper-level problem. For example, the MuJoCo simulator \cite{todorov2012mujoco} has been employed in an implicit bi-level approach for whole-body model predictive control of a humanoid robot \cite{koenemann2015whole}. The lower-level simulator solves a convex optimization problem in order to compute contact forces for rigid-body dynamics, and the results are utilized to perform rollouts and return finite-difference gradients to the upper-level iLQR optimizer. In contrast, explicit approaches have directly encoded the linear-complementarity-problem contact dynamics as constraints in a direct trajectory optimization method \cite{yunt2006trajectory, posa2014direct}. A related approach formulates contact dynamics as lower-level holonomic constraints \cite{mastalli2020crocoddyl}. Implicit integrators, which are a special-case of optimization-based dynamics and are widely used in collocation methods \cite{stryk1993numerical}, have been explored with differentiable dynamic programming generally \cite{jallet2022implicit} and specifically for models that require cloth \cite{zimmermann2021dynamic} or contact \cite{chatzinikolaidis2021trajectory} simulation. Implicit dynamics have also been trained to represent non-smooth dynamics \cite{pfrommer2021contactnets}, although this work has not been utilized for trajectory optimization. Direct methods with implicit lower-level problems have been used in locomotion applications for tracking reference trajectories with model predictive control \cite{lecleach2021fast} and a semi-direct method utilizes a lower-level friction problem for planning through contact events \cite{landry2019bilevel}. A related, sequential operator splitting framework is proposed in \cite{sindhwani2017sequential} and a derivative-free method that generates gradients via randomized smoothing for iLQR is also proposed \cite{suh2022bundled}. In this work we focus on implicit lower-level problems that can include cone constraints and indirect methods, specifically iLQR, as the upper-level optimizer.

	\section{Background}\label{background}
	In this section we provide background on the iLQR algorithm, implicit differentiation, and implicit integrators.
	
	\subsection{Iterative LQR}
	iLQR is an algorithm for trajectory optimization that solves instances of the following problem:
	\begin{equation}
	\begin{array}{ll}
	\underset{u_{1:T-1}}{\mbox{minimize}} & c_T(x_T) + \sum \limits_{t = 1}^{T-1} c_t(x_t, u_t) \label{ddp}\\
	\mbox{subject to} & x_{t+1} = f_t(x_t,u_t), \quad t = 1,\dots,T-1,\\
	& (x_1~\mbox{given}).
	\end{array}
	\end{equation}
	For a system with state $x_t \in \mathbf{R}^{n}$, control inputs $u_t \in \mathbf{R}^{m}$, time index $t$, initial state $x_1$, and discrete-time dynamics $f_t : \mathbf{R}^{n} \times \mathbf{R}^{m} \rightarrow \mathbf{R}^{n}$, the algorithm aims to minimize an objective with stage-cost functions, $c_t: \mathbf{R}^{n} \times \mathbf{R}^{m} \rightarrow \mathbf{R}$, and terminal-cost function, $c_T: \mathbf{R}^{n} \rightarrow \mathbf{R}$, over a planning horizon $T$.
	
	The algorithm utilizes gradients of the objective and dynamics, and exploits the problem's temporal structure. The state trajectory is defined implicitly by the initial state $x_1$ and only the control trajectory $u_{1:T-1}$ is parameterized as decision variables. Closed-loop rollouts enable this indirect method to work well in practice. The overall complexity is linear in the planning horizon and cubic in the control-input dimension \cite{tassa2007receding}.
	
	While efficient, the algorithm does not natively handle additional general equality or inequality constraints. This limitation has led to many variations of the classic algorithm in order to accommodate constraints in some capacity at the solver level. Box constraints have been incorporated at each step in the backward pass in order to enforce control limits \cite{tassa2014control}. An alternative approach embeds controls in barrier functions which smoothly approximate constraints \cite{marti2020squash}. Augmented Lagrangian methods and constrained backward and forward passes have also been proposed for handling general constraints \cite{howell2019altro, lantoine2012hybrid}.
	
	\subsection{Implicit Differentiation}
	An implicit function, $r : \mathbf{R}^n \times \mathbf{R}^p \rightarrow \mathbf{R}^n$, is defined as:
	\begin{equation}
	r(z^*; \theta) = 0, \label{implicit_function}
	\end{equation}
	for solutions $z^* \in \mathbf{R}^n$ and problem data $\theta \in \mathbf{R}^p$.
	
	At an equilibrium point, $z^*(\theta)$, the sensitivities of the solution with respect to the problem data, i.e., $\partial z / \partial \theta$, can be computed under certain conditions using the implicit function theorem \cite{dini1907lezioni}. First, we approximate (\ref{implicit_function}) to first order:
	\begin{equation}
	\frac{\partial r}{\partial z} \delta z + \frac{\partial r}{\partial \theta} \delta \theta = 0,
	\end{equation}
	and then the sensitivity of the solution is computed as:
	\begin{equation}
	\frac{\partial z}{\partial \theta} = -\Big(\frac{\partial r}{\partial z}\Big)^{-1} \frac{\partial r}{\partial \theta}. \label{solution_sensitivity}
	\end{equation}
	In the case that $\partial r / \partial z$ is not full rank, an approximate solution, e.g., least-squares, can be computed.
	
	\subsection{Implicit Integrators}    
	Unlike explicit integrators, i.e., $x_{t+1} = f_t(x_t, u_t)$, implicit integrators are an implicit function of the next state \cite{brudigam2020linear, manchester2016quaternion}, which cannot be separated from the current state and control input:
	\begin{equation}
		f_t(x_{t+1}, x_t, u_t) = 0, \label{implicit_dynamics}
	\end{equation}
	and are often used for numerical simulation of stiff systems due to improved stability and accuracy compared to explicit integrators, at the expense of increased computation \cite{wanner1996solving}.
	
	For direct trajectory optimization methods that parameterize states and controls and allow for dynamic infeasibility during iterates, such integrators are easily utilized since the state at the next time step is already available as a decision variable to the optimizer \cite{stryk1993numerical}. However, for indirect methods, like iLQR, that enforce strict dynamic feasibility at each iteration via rollouts, evaluating (\ref{implicit_dynamics}) requires a numerical solver to find a solution that satisfies this implicit function for given problem data.
	
	In practice, the next state can be found efficiently and reliably using Newton's method. Typically, the current state is used to initialize the solver and less than 10 iterations are required to solve the root-finding problem to machine precision.
	
	Having satisfied (\ref{implicit_dynamics}) to find the next state, we can compute the dynamics Jacobians using the implicit-function theorem. First, the dynamics are approximated to first order:
	\begin{equation}
		\frac{\partial f_t}{\partial x_{t+1}} \delta x_{t+1} + \frac{\partial f_t}{\partial x_t} \delta x_t + \frac{\partial f_t}{\partial u_t} \delta u_t  = 0,
	\end{equation}
	and then we solve for $\delta x_{t+1}$:
	\begin{equation}
		\delta x_{t+1} = -\Big(\frac{\partial f_t}{\partial x_{t+1}}\Big)^{-1} \Big(\frac{\partial f_t}{\partial x_t} \delta x_t
		+ \frac{\partial f_t}{\partial u_t} \delta u_t \Big).
	\end{equation}
	The Jacobians:
	\begin{align}
		\frac{\partial x_{t+1}}{\partial x_t} &= -\Big(\frac{\partial f_t}{\partial x_{t+1}}\Big)^{-1} \frac{\partial f_t}{\partial x_t}, \\
		\frac{\partial x_{t+1}}{\partial u_t} &= -\Big(\frac{\partial f_t}{\partial x_{t+1}}\Big)^{-1} \frac{\partial f_t}{\partial u_t}, \label{implicit_integrator_jacobians}
	\end{align}
	are returned at each time step.
	
	\section{Optimization-Based Dynamics} \label{id}
	In the previous section, we presented dynamics with implicit integrators that can be evaluated during rollouts and differentiated. In this section, we present a more general representation: optimization-based dynamics, which solve a constrained optimization problem in order to evaluate dynamics and use implicit differentiation to compute gradients by differentiating through the problem's optimality conditions.

	\subsection{Problem Formulation}
	For dynamics we require: fast and reliable evaluation, gradients that are useful to an upper-level optimizer, and (ideally) tight constraint satisfaction. We consider problems of the form:
	\begin{align}
	x_{t+1} \in z^*(\theta) = \underset{z \in \mathcal{K} \:| \: c(z; \theta) = 0 }{\mbox{arg min}} \ell(z; \theta)
	\label{argmin}
	\end{align}
	with decision variable $z \in \mathbf{R}^k$, problem data $\theta \in \mathbf{R}^p$, objective $\ell : \mathbf{R}^k \times \mathbf{R}^p \rightarrow \mathbf{R}$, equality constraints $c : \mathbf{R}^k \times \mathbf{R}^p \rightarrow \mathbf{R}^q$, and cone $\mathcal{K}$, which compactly represents combinations of free, positive-orthant, and second-order-cone constraints. Many problems from robotics can be specified in this form. For example, the objective could be the Lagrangian for a system, the cone constraints can represent joint limits, the problem data might comprise the current state and control, $\theta = (x_t, u_t)$, and the next state belongs to the optimal solution set. 
	
	\begin{algorithm}[t]
		\caption{Differentiable Interior-Point Method}\label{ip_algo}
		\begin{algorithmic}[1]
			\Procedure{Optimize}{$z, \theta$}
			\State \textbf{Parameters}: $\beta = 0.5, \gamma = 0.1$,
			\State \indent $\epsilon_{\mu} = 10^{-4}, \epsilon_{r} = 10^{-8}$
			\State \textbf{Initialize}: $z \in \mathcal{K}$, $\lambda = 0, \nu \in \mathcal{K}^*, \mu = 1.0, w_{\mu} = \{\}$
			\State $\bar{r} = r(w; \theta, \mu)$
			\State \textbf{Until} $\mu < \epsilon_{\mu}$ \textbf{do}
			\State \indent $\Delta w = (\Delta z, \Delta \lambda, \Delta \nu) = (\partial r / \partial w)^{-1} \bar{r}$
			\State \indent $\alpha \leftarrow 1$
			\State \indent \textbf{Until} $z - \alpha \Delta z \in \mathcal{K}$, $\nu - \alpha \Delta \nu \in \mathcal{K}^*$ \textbf{do}
			\State \indent \indent $\alpha \leftarrow \beta \alpha$
			\State \indent $\bar{r}_{+} = r(w-\alpha\Delta w; \theta, \mu)$
			\State \indent \textbf{Until} $\|\bar{r}_{+}\| < \|\bar{r}\|$ \textbf{do}
			\State \indent \indent $\alpha \leftarrow \beta \alpha$
			\State \indent \indent $\bar{r}_{+} = r(w-\alpha\Delta w; \theta, \mu)$
			\State \indent $w \leftarrow w - \alpha \Delta w$
			\State \indent $\bar{r} \leftarrow \bar{r}_{+}$
			\State \indent \textbf{If} $\|\bar{r}\| < \epsilon_{r}$ \textbf{do}
			\State \indent \indent $w_{\mu} \leftarrow w_{\mu} \cup w$
			\State \indent \indent $\mu \leftarrow \gamma \mu$
			\State $\partial w / \partial \theta \leftarrow \textbf{Differentiate}(w_{\mu}, \theta)$ \Comment{Eq. \ref{solution_sensitivity}}
			\State \textbf{Return} $w, \partial w / \partial \theta$
			\EndProcedure
		\end{algorithmic}
	\end{algorithm}

	\subsection{Interior-Point Method}
	One of the primary challenges in optimizing (\ref{argmin}) is selecting a solver that is well-suited to the requirements imposed by dynamics representations. Solvers for this class of problem are generally categorized as first-order projection \cite{stellato2020osqp, o2016conic, garstka2019cosmo} or second-order interior-point methods \cite{domahidi2013ecos, vandenberghe2010cvxopt}. The first approach optimizes (\ref{argmin}) by splitting the problem and alternating between inexpensive first-order methods, and potentially non-smooth, projections onto the cone. The second approach formulates and optimizes barrier subproblems, using second-order methods, along a central path, eventually converging to the cone's boundary in the limit \cite{boyd2004convex}. Second-order semi-smooth methods also exist \cite{ali2017semismooth}.
	
	The first-order projection-based methods, while fast, often can only achieve coarse constraint satisfaction and, importantly, the gradients returned are usually subgradients, which are less useful to an optimizer. In contrast, interior-point methods exhibit fast convergence, can achieve tight constraint satisfaction \cite{nocedal2006numerical}, and can return smooth gradients using a relaxed central-path parameter.
	
	Based on these characteristics, we utilize an interior-point method \cite{nocedal2006numerical} to optimize (\ref{argmin}). The idea behind this approach is that the cone constraints are handled using a logarithmic barrier and a sequence of relaxed, and easier to solve subproblems, optimized using Newton's method, converges to a solution of the original problem. 
	
	The optimality conditions for a barrier subproblem are:
	\begin{align}
		\partial \ell(z; \theta) / \partial z + (\partial c(z; \theta) / \partial z)^T \lambda - \nu &= 0, \label{opt_cond_lag}\\
		c(z; \theta) &= 0, \label{opt_cond_eq} \\
		z \circ \nu &= \mu \mathbf{e}, \label{opt_cond_comp} \\
		z \in \mathcal{K}, \, \nu &\in \mathcal{K}^*, \label{opt_cond_ineq}
	\end{align}
	with duals $\lambda \in \mathbf{R}^q$ and $\nu \in \mathbf{R}^k$, cone product denoted with the $\circ$ operator, central-path parameter $\mu \in \mathbf{R}_{+}$, and dual cone $\mathcal{K}^*$ \cite{domahidi2013ecos}. We consider free, nonnegative, and second-order cones.
	
	\begin{figure}[t]
		\centering
		\includegraphics[width=.225\textwidth]{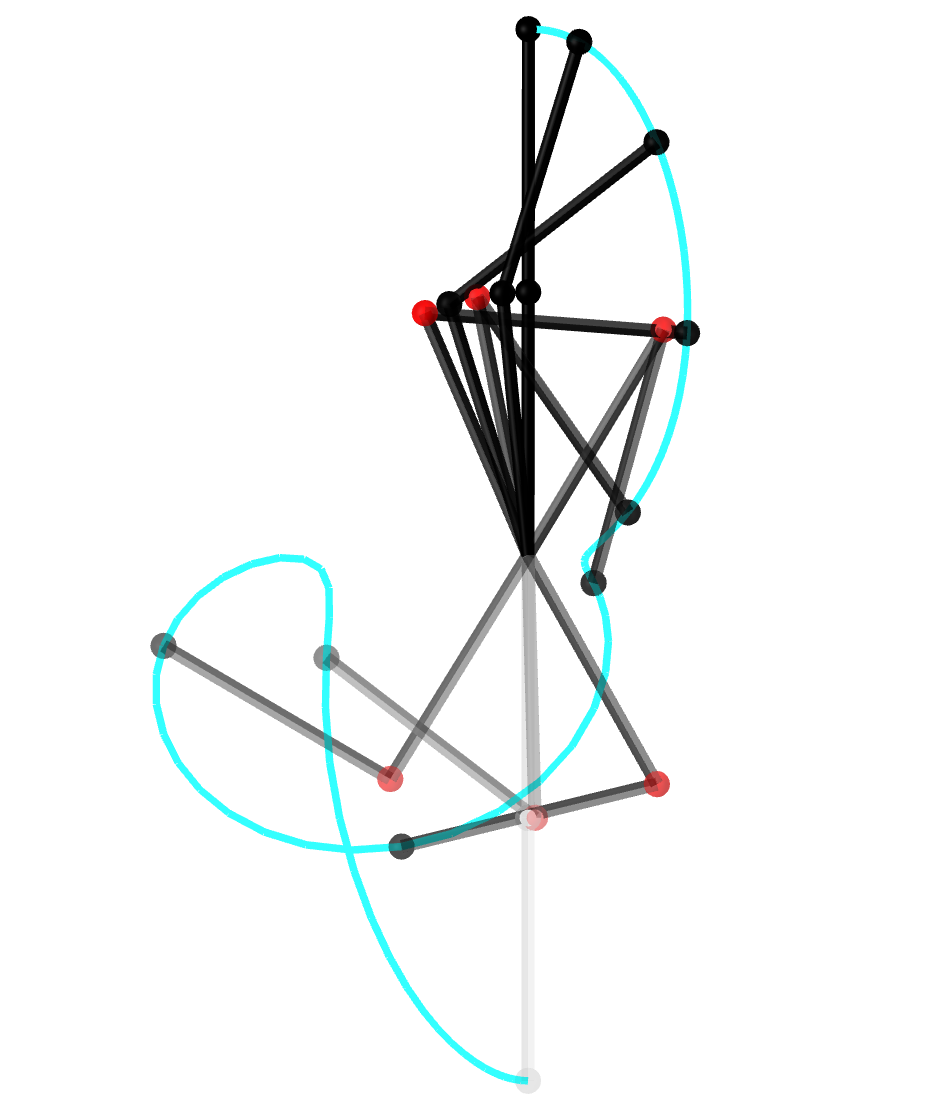}
		\includegraphics[width=.225\textwidth]{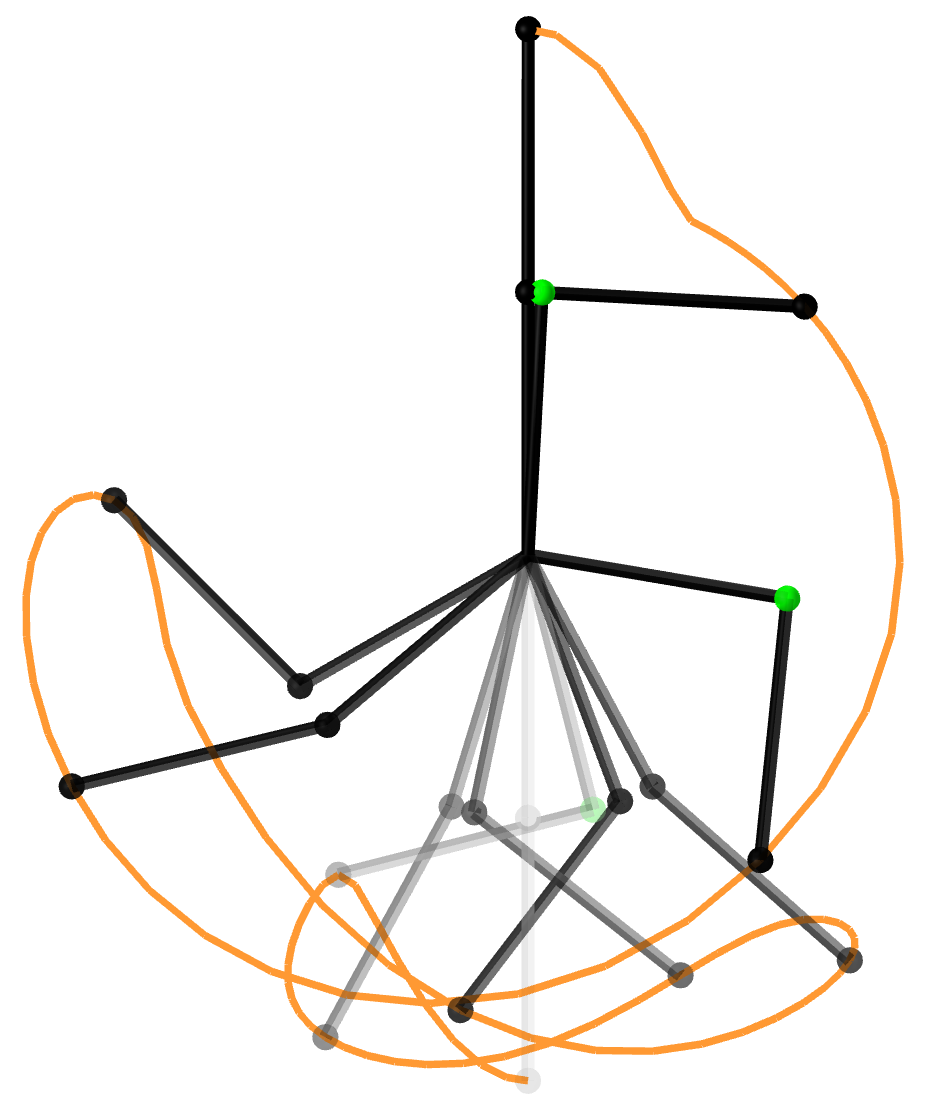}
		\caption[Comparison of acrobot swing-up behaviors]{Optimized trajectories for acrobot systems without (left) and with (right) joint limits performing a swing-up. Red and green elbow colors indicate violation or non-violation of the joint limits, respectively.}
		\label{acrobot_trajectories}
	\end{figure}
	
	To find a stationary point of (\ref{opt_cond_lag}-\ref{opt_cond_ineq}), for a fixed central-path parameter $\mu$, we consider $w = (z, \lambda, \nu) \in \mathbf{R}^{k + q + k}$ and a residual $r : \mathbf{R}^{k + q + k} \times \mathbf{R}^p \times \mathbf{R}_{+} \rightarrow \mathbf{R}^{k + p + k}$ comprising (\ref{opt_cond_lag}-\ref{opt_cond_comp}). A search direction, $\Delta w \in \mathbf{R}^{k + q + k}$, is computed using the residual and its Jacobian with respect to the decision variables at the current point. A backtracking line search is performed to ensure that a candidate step respects the cone constraints and reduces the norm of the residual. Once the residual norm is below a desired tolerance, we cache the current solution $w_{\mu}$, the central-path parameter is decreased, and the procedure is repeated until the central-path parameter is below a desired tolerance. We refer to \cite{boyd2004convex, nocedal2006numerical} for additional background on these methods.
	
	To differentiate (\ref{argmin}), we apply the implicit-function theorem (\ref{solution_sensitivity}) to the residual at an intermediate solution, $w_{\mu}$. In practice, we find that performing implicit differentiation with a solution point having a relaxed central-path parameter returns smooth gradients that improve the convergence behavior of the upper-level optimizer. 
	
	\begin{table}[t]
		\centering
		\caption[Numerical results for acrobot swing-up]{Comparison of objective, iterations, goal constraint violation, and solve time between MuJoCo and optimization-based dynamics (OD) with nominal and joint-limited models for acrobot. When limited, the MuJoCo model violates the joint limits and fails at the swing-up task.}
		\begin{tabular}{c c c c c}
			\toprule
			&
			\textbf{MuJoCo} &
			\textbf{MuJoCo (limited)} &
			\textbf{OD} &
			\textbf{OD (limited)} \\
			\toprule
			objective & 395.6 & 1.9e{6} & 48.9 & 84.2 \\
			iterations & 207  & 95 & 256 & 907 \\
			violation & 3e{-}4 & 0.2 & 5e{-}4 & 1e{-}3\\
			time (s) & 0.9 & 1.0 & 0.5 & 6.5\\
			\toprule
		\end{tabular}
		\label{acrobot_results}
	\end{table}

	\section{Examples}\label{examples}
	We formulate optimization-based dynamics models and use iLQR to perform bi-level trajectory optimization for a number of examples that highlight how these representations can be constructed and demonstrate that trajectories for non-smooth and constrained dynamics can be optimized. Additionally, we provide a comparison of our approach with MuJoCo using finite-difference gradients, contact-implicit trajectory optimization, and gradients generated via randomized smoothing.
	
	Throughout, we use implicit midpoint integrators, quadratic costs, and for convenience, employ an augmented Lagrangian method to enforce any remaining trajectory-level constraints (e.g., terminal constraints), not handled implicitly by the dynamics representation. Our implementation, models, and all of the experiments are available here:
	\begin{center}
		\url{https://github.com/thowell/optimization_dynamics}.
	\end{center}
	
	\subsection{Acrobot with Joint Limits}
	
	We model an acrobot \cite{tedrake2014underactuated} with joint limits on the actuated elbow. These limits are enforced with a signed-distance constraint:
	\begin{equation}
	\phi(q) = \begin{bmatrix}\pi / 2 - q_{\mbox{e}} \\ q_{\mbox{e}} + \pi / 2 \end{bmatrix} \geq 0, \label{joint_limits}
	\end{equation}
	where $q \in \mathbf{R}^2$ is the system's configuration and $q_{\mbox{e}}$ is the elbow angle. Additional constraints:
	\begin{equation}
	\lambda \circ \phi(q) = 0, \quad \lambda \geq 0, \label{impact_con}
	\end{equation}
	enforce physical behaviors that impact impulses $\lambda_t \in \mathbf{R}^2$ can only be non-negative, i.e., repulsive not attractive, and that they are only applied to the system
	when the joint reaches a limit. Relaxing the complementarity constraint via a central-path parameter, introducing a slack variable for the signed-distance function $\phi$, and combining this reformulation with the system's dynamics results in a problem formulation (\ref{opt_cond_lag}-\ref{opt_cond_ineq}) that can be optimized with Algorithm \ref{ip_algo}.
	
	Unlike approaches that include joint limits at the solver level, including the impact (\ref{impact_con}) and limit (\ref{joint_limits}) constraints at the dynamics level enables impact forces encountered at joint stops to be optimized and applied to the system.
	
	The system has $n = 4$ states and $m = 1$ controls. We plan a swing-up trajectory over a horizon $T = 101$ with a time step $h = 0.05$. The optimizer is initialized with random controls.
	
	We compare unconstrained and joint-limited systems, the optimized motions are shown in Fig. \ref{acrobot_trajectories}. The unconstrained system violates joint limits, while the joint-limited system reaches the limits without exceeding them and finds a motion utilizing additional pumps. Additionally, we compare our optimization-based dynamics with the MuJoCo physics simulator and gradients provided via finite-differencing. For the unconstrained system, the optimizer is able to successfully find a swing-up trajectory. For the joint-limited system, the trajectory optimizer fails. We also note that MuJoCo is unable to enforce hard joint limits and that such constraints typically require tuning the solver parameters to produce realistic looking behavior. The results are summarized in Table \ref{acrobot_results}.
	
	\begin{figure}[t]
		\centering
		\includegraphics[width=.3\textwidth]{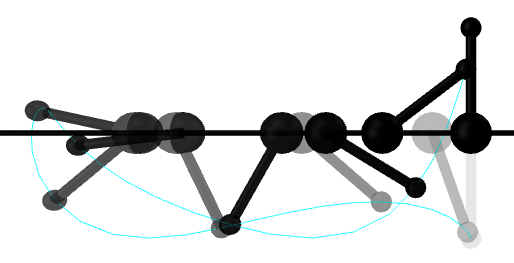}
		\includegraphics[width=.3\textwidth]{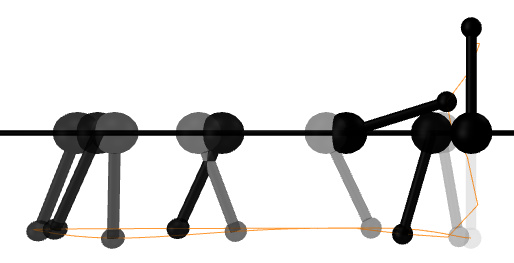}
		\caption{Optimized trajectories for cart-pole performing a swing-up without (top, friction coefficient $=0$) and with (bottom, friction coefficient $=0.35$) joint friction.}
		\label{cartpole_state}
	\end{figure}
	
	\subsection{Cart-Pole with Coulomb Friction}
	
	A cart-pole \cite{tedrake2014underactuated} is modeled that experiences Coulomb friction \cite{moreau2011unilateral} on both its slider and pendulum arm. The friction force that maximally dissipates kinetic energy is the solution to the following optimization problem:
	\begin{equation}
	\begin{array}{ll}
	\underset{b}{\mbox{minimize}} & v^T b\\
	\mbox{subject to} &\|b\|_2 \leq N, \\
	\end{array} \label{mdp}
	\end{equation}
	where $v \in \mathbf{R}^{d-1}$ is the joint velocity, $b \in \mathbf{R}^{d-1}$ is the friction force, and $ N \in \mathbf{R}_{+}$ is the friction-cone limit \cite{lobo1998applications}. The optimality conditions for the cone program's barrier subproblem are:
	\begin{align}
	\begin{bmatrix} y^T & v^T \end{bmatrix}^T - \eta &= 0, \label{friction_stationarity} \\
	\beta_{(1)} - N &= 0, \label{friction_normal_eq}\\
	\beta \circ \eta &= \mu \mathbf{e} \label{friction_comp},\\
	\beta, \eta &\in \mathcal{Q}^d \label{friction_cones},
	\end{align}
	with $\beta \in \mathbf{R}^d$, such that $b = \beta_{(2:d)}$, and dual variables $y \in \mathbf{R}$ and $\eta \in \mathbf{R}^d$ for the equality and cone constraints, respectively.
	
	The system has $n = 4$ states, $m = 1$ controls, and dimension $d = 2$. We make the modeling assumption that $N$, which is a function of the friction coefficient, is fixed for both joints. We plan a swing-up trajectory for a horizon $T = 51$ and time step $h = 0.05$. The optimizer is initialized with an impulse at the first time step and zeros for the remaining initial control inputs.
	
	We compare the trajectories optimized for systems with increasing amounts of friction, i.e., increasing $N$, in Fig. \ref{cartpole_state}. In the presence of friction, the system performs a more aggressive manuever at the end of the trajectory in order to achieve the swing-up. The results are summarized in Table \ref{cartpole_results}.
	
	\begin{table}[t]
		\centering
		\caption[Numerical results for cart-pole swing-up with friction]{Comparison of objective, iterations, goal constraint violation, and solve time for cart-pole model with different friction coefficients.}
		\begin{tabular}{c c c c c c}
			\toprule
			\textbf{Friction} &
			\textbf{0.0} &
			\textbf{0.01} &
			\textbf{0.1} &
			\textbf{0.25} & 
			\textbf{0.35} \\
			\toprule
			objective & 5.0 & 5.5 & 11.6 & 76.3 & 163.1\\
			iterations & 456 & 485 & 406 & 472 & 943\\
			violation & 1e{-}4 & 5e{-}4 & 6e{-}4 & 1e{-}3 & 5e{-}3 \\
			time (s) & 0.36 & 2.8 & 3.1 & 4.6 & 9.8\\
			\toprule
		\end{tabular}
		\label{cartpole_results}
	\end{table}
	
	\subsection{Hopper Gait}
	
	We generate a gait for a Raibert hopper \cite{raibert1989dynamically}. The system's dynamics are modeled as a nonlinear complementarity problem \cite{lecleach2021fast}. The 2D system has four generalized coordinates: lateral and vertical body positions, body orientation, and leg length. There are $m = 2$ control inputs: a body moment and leg force. The state is $n = 8$, comprising the system's current and previous configurations. The  horizon is $T = 21$ with time step $h = 0.05$. The initial state is optimized and a trajectory-level periodicity constraint is employed to ensure that the first and final states, with the exception of the lateral positions, are equivalent. Finally, we initialize the solver with controls that maintain the system in an upright configuration.
	
	We compare our implicit bi-level approach to a direct method for contact-implicit trajectory optimization \cite{manchester2020variational} that transcribes the problem and solves it with Ipopt \cite{wachter2006implementation}. We use the same models, cost functions, and comparable optimizer parameters and tolerances. 
	
	The complexity of the classic iLQR algorithm is: $O(T m^3)$ \cite{tassa2007receding}, while a direct method for trajectory optimization that exploits the problem's temporal structure is: $O\Big(T (n^3 + n^2 m) \Big)$ \cite{wang2009fast}. The interior-point method for the dynamics generally has complexity: $O(a^3)$, although specialized solvers can reduce this cost, where $a$ is the number of primal and dual variables \cite{nocedal2006numerical}. The complexity of optimization-based dynamics and iLQR is typically dominated by the cost of the interior-point method, which is incurred at each time step in the planning horizon, so the overall complexity is $O(T a^3)$. The hopper problem has $a = 20$ and $n = 8$ and $m = 2$ with our approach. For the contact-implicit approach, the controls are augmented with contact variables, increasing the dimension to $m = 17$ at each time step. As a result, contact-implicit trajectory optimization has lower complexity in the case where a structure exploiting direct method is employed, although this is not necessarily the case when a general-purpose solver like Ipopt is called.
	
\begin{figure}[t]
		\centering
		\includegraphics[width=.5\textwidth]{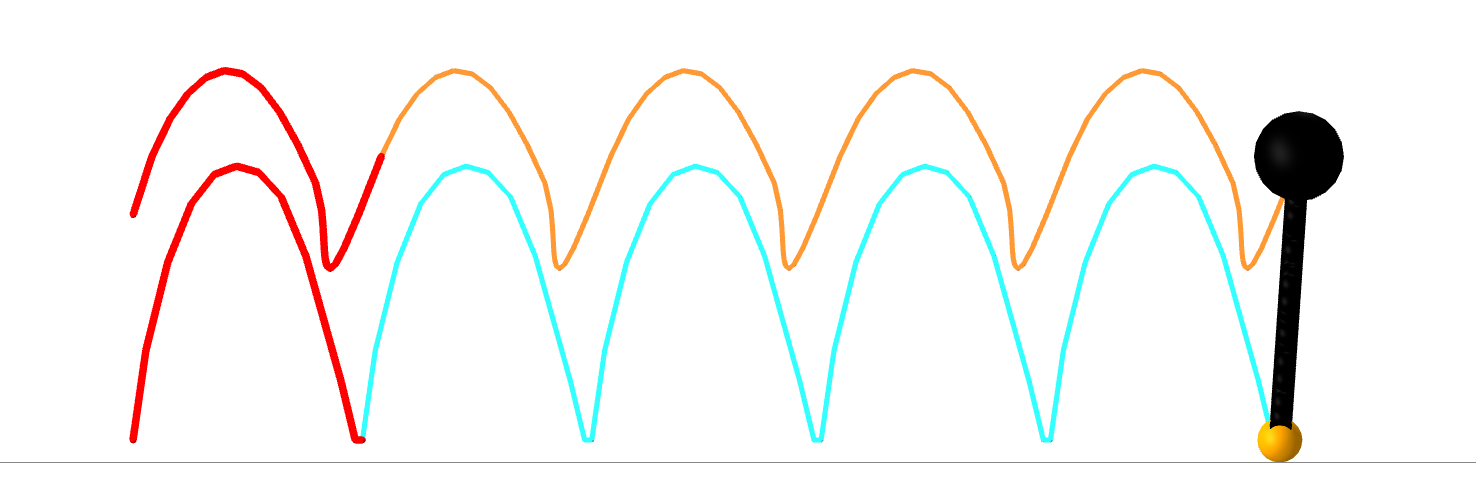}
		\caption{Hopper gait. Optimized template (red) for the system's body (orange) and foot (blue) trajectories is repeated to form a gait.}
		\label{hopper_gait}
	\end{figure}

	An optimized gait is shown in Fig. \ref{hopper_gait}. We find that our implicit approach requires fewer, but more expensive iterations, while contact-implicit trajectory optimization requires more, but less expensive iterations. Additionally, our approach more consistently returns good trajectories. Numerical comparison results are provided in Table \ref{hopper_bilevel_comparison}.
	
	\subsection{Rocket Landing}
	
	We plan a soft-landing for a rocket with 6 degrees of freedom that must transition from a horizontal to vertical orientation prior to landing while respecting thrust constraints. Unlike prior work that enforces such constraints at the optimizer level \cite{blackmore2010minimum}, we instead enforce these constraints at the dynamics level. This enables the optimizer to provide smooth controls to the dynamics, which then internally constrain the input to satisfy the system's limits at every iteration.
	
	The cone projection problem, which finds the thrust vector that is closest to the optimizer's input thrust while satisfying constraints, is formulated as:
	\begin{equation}
		\begin{array}{ll}
			\underset{u}{\mbox{minimize}} & \frac{1}{2} \|u - \bar{u}\|_2^2 \\
			\mbox{subject to} & \|u_{2:3}\|_2 \leq u_1, \\
			& u_{\mbox{min}} \leq u_1 \leq u_{\mbox{max}},
		\end{array} \label{cone_projection}
	\end{equation}
	where $u, \bar{u} \in \mathbf{R}^3$ are the optimized and provided thrust vectors, respectively, $u_1$ is the component of thrust along the longitudinal axis of the rocket, $u_{\mbox{min}}$ and $u_{\mbox{max}}$ are limits on this value.
	
	The system has $n = 12$ states and $m = 3$ controls that are first projected using (\ref{cone_projection}) before being applied to dynamics. The planning horizon is $T = 61$ and time step is $h = 0.05$. The controls are initialized with small random values, the initial pose is offset from the target, and the rocket has initial downward velocity. A constraint enforces the rocket's final pose, a zero velocity, vertical-orientation configuration in a landing zone.
	
\begin{figure}[t]
		\centering
		\includegraphics[width=0.3\textwidth]{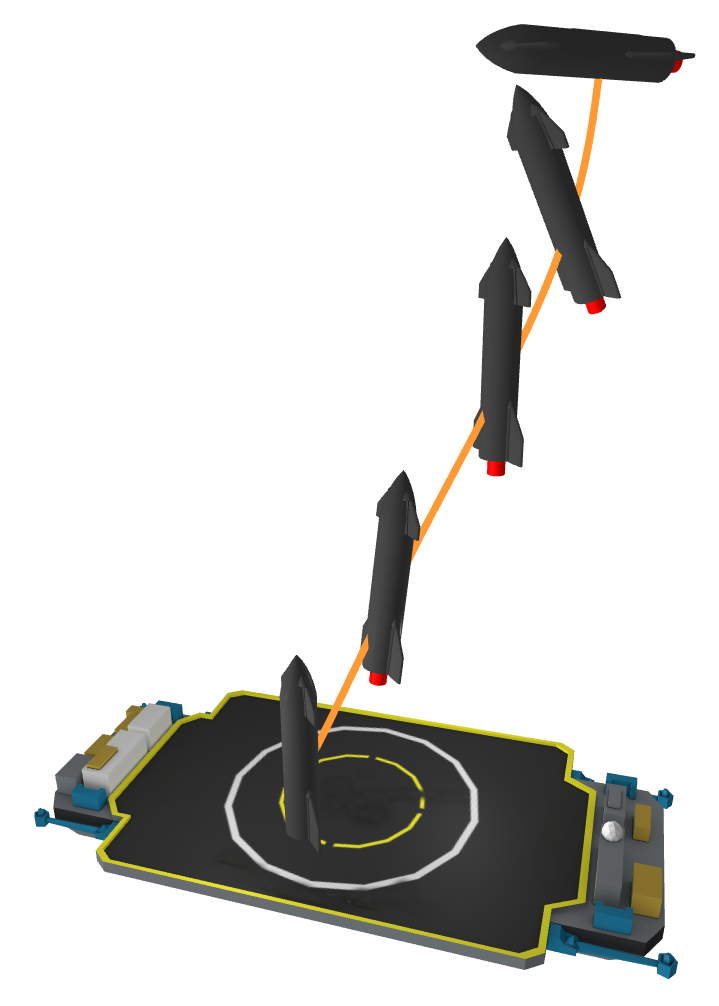}
		\caption{Position trajectory (orange) for rocket performing soft landing. The system first reorients from a horizontal to vertical pose before landing with zero velocity in a target zone while respecting thrust constraints.}
		\label{rocket_ghost}
	\end{figure}

	The optimizer requires 547 iterations and takes 8.1 seconds to find a plan that successfully reorients the rocket prior to landing while respecting the thrust constraints. Without the cone constraint, the optimizer requires 521 iterations and 1.9 seconds to find a solution. However, the thrust cone constraint is violated at the first time steps. The optimized position trajectory is shown in Fig. \ref{rocket_ghost}.
	
	\subsection{Planar Push}
	For the canonical manipulation problem of planar pushing \cite{hogan2016feedback}, we optimize the positions and forces of a robotic manipulator's circular end-effector in order to slide a planar block into a reoriented goal pose (Fig.~\ref{planar_push}).
	
	The system's end-effector is modeled as a fully actuated particle in 2D that can move the block (with 2D translation and orientation) via impact and friction while the two systems are in contact. The block is modeled with point friction at each of its four corners and a signed-distance function, modeled as a $p$-norm with $p=10$, is employed that enables the end-effector to push on any of the block's sides.
	
	The system has $n = 10$ states, $m = 2$ controls, the planning horizon is $T = 26$, and the time step is $h = 0.1$. We initialize the end-effector with a control input that overcomes stiction to move the block. The block is initialized at the origin with the pusher in contact on its left side, and the block is constrained at the trajectory-level to reach a goal pose.
	
	\begin{table}[t]
		\centering
		\caption[Numerical results for hopper gait]{Comparison between contact-implicit trajectory optimization (CI) and optimization-based dynamics + iLQR (OD) for hopper-gait examples. The objective, iterations, gait constraint violation, and solve times are compared.}
		\begin{tabular}{c c c c c c c}
			\toprule
			&
			\textbf{CI (1)} &
			\textbf{CI (2)} &
			\textbf{CI (3)} &
			\textbf{OD (1)} &
			\textbf{OD (2)} &
			\textbf{OD (3)} 
			\\
			\toprule
			objective & 3.5 & 20.4 & 2.9 & 3.5 & \textbf{19.8} & \textbf{2.4} \\
			iterations & 122 & 114 & 107 & \textbf{38} & \textbf{47} & \textbf{25} \\
			violation & \textbf{1\text{e-}9} & \textbf{1\text{e-}9} & \textbf{1\text{e-}9} & 3\text{e-}4 & 3\text{e-}4 & 9e{-}4\\
			time (s) & 2.0 & 2.0 & 1.8 & \textbf{0.3} & \textbf{0.4} & \textbf{0.3}\\
			\toprule
		\end{tabular}
		\label{hopper_bilevel_comparison}
	\end{table}

	We compare the smooth gradients returned by differentiating through our interior-point solver with randomized smoothing to generate a zero-order gradient bundle \cite{suh2022bundled}. We use $100$ samples and noise sampled from a zero-mean unit Gaussian with scaling $\sigma=1.0e\mbox{-}4$. Ultimately, we find that both approaches result in similar convergence behavior of the upper-level optimizer. However, the gradient-bundle approach necessitates parallel computation of dynamics evaluations (which we do not explore in our comparison) in order to be a tractable approach, whereas the implicit gradients are much more efficient in a serial-computational setting. Results are provided in Table \ref{planarpush_results}.
	
	\section{Conclusion}\label{conclusion}
	
	In this work we have presented a bi-level optimization framework that utilizes lower-level optimization-based dynamics for planning. This representation enables expressive dynamics by including constraint handling, internal states, and additional physical behavior modeling at the dynamics level, abstracted away from the upper-level optimizer, enabling classically unconstrained solvers to optimize motions for more complex systems.

	\emph{Limitations:} One of the primary drawbacks to this approach is the lack of convergence guarantees for finding a solution that satisfies the dynamics representation. In practice, we find that the solver converges reliably. However, there are cases where the solver fails to meet specified tolerances. In this event we have the optimizer return the current solution and its sensitivities. We find that occasional failures like this often do not impair overall planning and that the optimizer can eventually find a dynamically feasible trajectory. Just as robust, large-scale solvers such as Ipopt \cite{wachter2006implementation} fallback to their restoration mode when numerical difficulties occur, our basic interior-point method is likely to be improved by including such a fallback routine, perhaps specific to a particular system. Additionally, we find that standard techniques such as: problem scaling, appropriate hyperparameter selection, and warm starting greatly improve the reliability of this approach.
	
	Other potential weaknesses are the increased serial nature of the approach and cost of evaluating the dynamics and their gradients. First, iLQR is a serial algorithm, dominated by forward rollouts and backward Riccati recursions that cannot be parallelized. Similarly, the interior-point solver is a serial method dominated by a matrix factorization and back substitution that is generally not amenable to parallelization either. Second, because an iterative solver is utilized to evaluate the dynamics, calls to the dynamics are inherently more expensive compared to an explicit dynamics representation. However, such overhead can potentially be mitigated with a problem-specific solver that can exploit specialized constraints or sparsity, unlike an off-the-shelf solver's generic routines.

	\emph{Future Work:} There are numerous avenues for future work exploring optimization-based dynamics for bi-level trajectory optimization. First, in this work we explore constrained optimization problems solved with a second-order method. Another interesting problem class to consider are energy-based models \cite{lecun2006tutorial}, potentially optimized with first-order Langevin dynamics \cite{schlick2010molecular}. Second, we find that returning gradients optimized with a relaxed central-path parameter greatly improves the convergence behavior of the upper-level optimizer. An analysis of, or method for, returning gradients from the lower-level problem that best aid an upper-level optimizer would be useful. Finally, this bi-level framework could be extended to a tri-level setting where the highest-level optimizer autotunes an objective to generate model predictive control policies in an imitation-learning setting.
	
	\begin{table}[t]
		\centering
		\caption[Gradient comparison between implicit gradients and gradient bundles for planar-push task]{Comparison between gradients generated as zero-order gradient bundles and via implicit differentiation for planar-push translation (T) and rotation (R) tasks. $^*$Gradient bundles are evaluated serially.}
		\begin{tabular}{c c c c c}
			\toprule
			&
			\textbf{Bundle (T)} &
			\textbf{Bundle (R)} &
			\textbf{Implicit (T)} &
			\textbf{Implicit (R)}
			\\
			\toprule
			iterations & 18 & \textbf{32} & \textbf{17} & 33 \\
			time (s) & $+100.0^*$ & $+100.0^*$ & \textbf{8.0} & \textbf{18.3} \\
			\toprule
		\end{tabular}
		\label{planarpush_results}
	\end{table}

	\bibliographystyle{IEEEtran}
	\bibliography{IEEEabrv,main}

\begin{thebibliography}{10}
\providecommand{\url}[1]{#1}
\csname url@samestyle\endcsname
\providecommand{\newblock}{\relax}
\providecommand{\bibinfo}[2]{#2}
\providecommand{\BIBentrySTDinterwordspacing}{\spaceskip=0pt\relax}
\providecommand{\BIBentryALTinterwordstretchfactor}{4}
\providecommand{\BIBentryALTinterwordspacing}{\spaceskip=\fontdimen2\font plus
\BIBentryALTinterwordstretchfactor\fontdimen3\font minus
  \fontdimen4\font\relax}
\providecommand{\BIBforeignlanguage}[2]{{%
\expandafter\ifx\csname l@#1\endcsname\relax
\typeout{** WARNING: IEEEtran.bst: No hyphenation pattern has been}%
\typeout{** loaded for the language `#1'. Using the pattern for}%
\typeout{** the default language instead.}%
\else
\language=\csname l@#1\endcsname
\fi
#2}}
\providecommand{\BIBdecl}{\relax}
\BIBdecl

\bibitem{jacobson1970differential}
D.~H. Jacobson and D.~Q. Mayne, \emph{Differential Dynamic Programming}.\hskip
  1em plus 0.5em minus 0.4em\relax Elsevier Publishing Company, 1970, no.~24.

\bibitem{sinha2017review}
A.~Sinha, P.~Malo, and K.~Deb, ``A review on bilevel optimization: {F}rom
  classical to evolutionary approaches and applications,'' \emph{IEEE
  Transactions on Evolutionary Computation}, vol.~22, no.~2, pp. 276--295,
  2017.

\bibitem{todorov2012mujoco}
E.~Todorov, T.~Erez, and Y.~Tassa, ``Mu{J}o{C}o: {A} physics engine for
  model-based control,'' in \emph{IEEE/RSJ International Conference on
  Intelligent Robots and Systems}, 2012, pp. 5026--5033.

\bibitem{koenemann2015whole}
J.~Koenemann, A.~Del~Prete, Y.~Tassa, E.~Todorov, O.~Stasse, M.~Bennewitz, and
  N.~Mansard, ``Whole-body model-predictive control applied to the {HRP}-2
  humanoid,'' in \emph{IEEE/RSJ International Conference on Intelligent Robots
  and Systems}, 2015, pp. 3346--3351.

\bibitem{yunt2006trajectory}
K.~Yunt and C.~Glocker, ``Trajectory optimization of mechanical hybrid systems
  using {SUMT},'' in \emph{IEEE International Workshop on Advanced Motion
  Control}, 2006, pp. 665--671.

\bibitem{posa2014direct}
M.~Posa, C.~Cantu, and R.~Tedrake, ``A direct method for trajectory
  optimization of rigid bodies through contact,'' \emph{The International
  Journal of Robotics Research}, vol.~33, no.~1, pp. 69--81, 2014.

\bibitem{mastalli2020crocoddyl}
C.~Mastalli, R.~Budhiraja, W.~Merkt, G.~Saurel, B.~Hammoud, M.~Naveau,
  J.~Carpentier, L.~Righetti, S.~Vijayakumar, and N.~Mansard, ``Crocoddyl: {A}n
  efficient and versatile framework for multi-contact optimal control,'' in
  \emph{IEEE International Conference on Robotics and Automation}, 2020, pp.
  2536--2542.

\bibitem{stryk1993numerical}
O.~Von~Stryk, ``Numerical solution of optimal control problems by direct
  collocation,'' in \emph{Optimal Control}.\hskip 1em plus 0.5em minus
  0.4em\relax Springer, 1993, pp. 129--143.

\bibitem{jallet2022implicit}
W.~Jallet, N.~Mansard, and J.~Carpentier, ``Implicit differential dynamic
  programming,'' in \emph{International Conference on Robotics and Automation},
  2022, pp. 1455--1461.

\bibitem{zimmermann2021dynamic}
S.~Zimmermann, R.~Poranne, and S.~Coros, ``Dynamic manipulation of deformable
  objects with implicit integration,'' \emph{IEEE Robotics and Automation
  Letters}, vol.~6, no.~2, pp. 4209--4216, 2021.

\bibitem{chatzinikolaidis2021trajectory}
I.~Chatzinikolaidis and Z.~Li, ``Trajectory optimization of contact-rich
  motions using implicit differential dynamic programming,'' \emph{IEEE
  Robotics and Automation Letters}, vol.~6, no.~2, pp. 2626--2633, 2021.

\bibitem{pfrommer2021contactnets}
S.~Pfrommer, M.~Halm, and M.~Posa, ``{C}ontact{N}ets: {L}earning discontinuous
  contact dynamics with smooth, implicit representations,'' in \emph{Conference
  on Robot Learning}, 2021, pp. 2279--2291.

\bibitem{lecleach2021fast}
S.~Le~Cleac'h, T.~A. Howell, S.~Yang, C.~Lee, J.~Zhang, A.~Bishop, M.~Schwager,
  and Z.~Manchester, ``Fast contact-implicit model-predictive control,''
  \emph{arXiv:2107.05616}, 2021.

\bibitem{landry2019bilevel}
B.~Landry, J.~Lorenzetti, Z.~Manchester, and M.~Pavone, ``Bilevel optimization
  for planning through contact: {A} semidirect method,'' in \emph{The
  International Symposium of Robotics Research}, 2019, pp. 789--804.

\bibitem{sindhwani2017sequential}
V.~Sindhwani, R.~Roelofs, and M.~Kalakrishnan, ``Sequential operator splitting
  for constrained nonlinear optimal control,'' in \emph{American Control
  Conference}, 2017, pp. 4864--4871.

\bibitem{suh2022bundled}
H.~J.~T. Suh, T.~Pang, and R.~Tedrake, ``Bundled gradients through contact via
  randomized smoothing,'' \emph{IEEE Robotics and Automation Letters}, vol.~7,
  no.~2, pp. 4000--4007, 2022.

\bibitem{tassa2007receding}
Y.~Tassa, T.~Erez, and W.~D. Smart, ``Receding horizon differential dynamic
  programming,'' \emph{{A}dvances in {N}eural {I}nformation {P}rocessing
  {S}ystems}, vol.~20, 2007.

\bibitem{tassa2014control}
Y.~Tassa, N.~Mansard, and E.~Todorov, ``Control-limited differential dynamic
  programming,'' in \emph{IEEE International Conference on Robotics and
  Automation}, 2014, pp. 1168--1175.

\bibitem{marti2020squash}
J.~Marti-Saumell, J.~Sol{\`a}, C.~Mastalli, and A.~Santamaria-Navarro,
  ``Squash-box feasibility driven differential dynamic programming,'' in
  \emph{International Conference on Intelligent Robots and Systems}.\hskip 1em
  plus 0.5em minus 0.4em\relax IEEE, 2020, pp. 7637--7644.

\bibitem{howell2019altro}
T.~A. Howell, B.~E. Jackson, and Z.~Manchester, ``{ALTRO}: {A} fast solver for
  constrained trajectory optimization,'' in \emph{IEEE/RSJ International
  Conference on Intelligent Robots and Systems}, 2019, pp. 7674--7679.

\bibitem{lantoine2012hybrid}
G.~Lantoine and R.~P. Russell, ``A hybrid differential dynamic programming
  algorithm for constrained optimal control problems. {P}art 1: {T}heory,''
  \emph{Journal of Optimization Theory and Applications}, vol. 154, no.~2, pp.
  382--417, 2012.

\bibitem{dini1907lezioni}
U.~Dini, \emph{Lezioni di analisi infinitesimale}.\hskip 1em plus 0.5em minus
  0.4em\relax Fratelli Nistri, 1907, vol.~1.

\bibitem{brudigam2020linear}
J.~Br{\"u}digam and Z.~Manchester, ``Linear-time variational integrators in
  maximal coordinates,'' in \emph{International Workshop on the Algorithmic
  Foundations of Robotics}, 2020, pp. 194--209.

\bibitem{manchester2016quaternion}
Z.~R. Manchester and M.~A. Peck, ``Quaternion variational integrators for
  spacecraft dynamics,'' \emph{Journal of Guidance, Control, and Dynamics},
  vol.~39, no.~1, pp. 69--76, 2016.

\bibitem{wanner1996solving}
G.~Wanner and E.~Hairer, \emph{Solving Ordinary Differential Equations
  {II}}.\hskip 1em plus 0.5em minus 0.4em\relax Springer Berlin Heidelberg,
  1996, vol. 375.

\bibitem{stellato2020osqp}
B.~Stellato, G.~Banjac, P.~Goulart, A.~Bemporad, and S.~Boyd, ``{OSQP}: {A}n
  operator splitting solver for quadratic programs,'' \emph{Mathematical
  Programming Computation}, vol.~12, no.~4, pp. 637--672, 2020.

\bibitem{o2016conic}
B.~O’Donoghue, E.~Chu, N.~Parikh, and S.~Boyd, ``Conic optimization via
  operator splitting and homogeneous self-dual embedding,'' \emph{Journal of
  Optimization Theory and Applications}, vol. 169, no.~3, pp. 1042--1068, 2016.

\bibitem{garstka2019cosmo}
M.~Garstka, M.~Cannon, and P.~Goulart, ``{COSMO}: {A} conic operator splitting
  method for large convex problems,'' in \emph{European Control Conference},
  2019, pp. 1951--1956.

\bibitem{domahidi2013ecos}
A.~Domahidi, E.~Chu, and S.~Boyd, ``{ECOS}: {A}n {SOCP} solver for embedded
  systems,'' in \emph{European Control Conference}, 2013, pp. 3071--3076.

\bibitem{vandenberghe2010cvxopt}
L.~Vandenberghe, ``The {C}{V}{X}{O}{P}{T} linear and quadratic cone program
  solvers,'' \emph{Online: http://cvxopt.org/documentation/coneprog.pdf}, 2010.

\bibitem{boyd2004convex}
S.~Boyd and L.~Vandenberghe, \emph{Convex Optimization}.\hskip 1em plus 0.5em
  minus 0.4em\relax Cambridge University Press, 2004.

\bibitem{ali2017semismooth}
A.~Ali, E.~Wong, and J.~Z. Kolter, ``A semismooth {N}ewton method for fast,
  generic convex programming,'' in \emph{International Conference on Machine
  Learning}, 2017, pp. 70--79.

\bibitem{nocedal2006numerical}
J.~Nocedal and S.~J. Wright, \emph{Numerical Optimization}, 2nd~ed.\hskip 1em
  plus 0.5em minus 0.4em\relax {Springer}, 2006.

\bibitem{tedrake2014underactuated}
R.~Tedrake, ``Underactuated robotics: {A}lgorithms for walking, running,
  swimming, flying, and manipulation (course notes for {MIT} 6.832),'' 2022.

\bibitem{moreau2011unilateral}
J.~J. Moreau, ``On unilateral constraints, friction and plasticity,'' in
  \emph{New Variational Techniques in Mathematical Physics}.\hskip 1em plus
  0.5em minus 0.4em\relax Springer, 2011, pp. 171--322.

\bibitem{lobo1998applications}
M.~S. Lobo, L.~Vandenberghe, S.~Boyd, and H.~Lebret, ``Applications of
  second-order cone programming,'' \emph{Linear {A}lgebra and its
  Applications}, vol. 284, no. 1-3, pp. 193--228, 1998.

\bibitem{raibert1989dynamically}
M.~H. Raibert, H.~B. Brown~Jr., M.~Chepponis, J.~Koechling, J.~K. Hodgins,
  D.~Dustman, W.~K. Brennan, D.~S. Barrett, C.~M. Thompson, J.~D. Hebert,
  W.~Lee, and B.~Lance, ``Dynamically stable legged locomotion,'' Massachusetts
  Institute of Technology Cambridge Artificial Intelligence Lab, Tech. Rep.,
  1989.

\bibitem{manchester2020variational}
Z.~Manchester and S.~Kuindersma, ``Variational contact-implicit trajectory
  optimization,'' in \emph{Robotics Research}.\hskip 1em plus 0.5em minus
  0.4em\relax Springer, 2020, pp. 985--1000.

\bibitem{wachter2006implementation}
A.~W{\"a}chter and L.~T. Biegler, ``On the implementation of an interior-point
  filter line-search algorithm for large-scale nonlinear programming,''
  \emph{Mathematical Programming}, vol. 106, no.~1, pp. 25--57, 2006.

\bibitem{wang2009fast}
Y.~Wang and S.~Boyd, ``Fast model predictive control using online
  optimization,'' \emph{IEEE Transactions on Control Systems Technology},
  vol.~18, no.~2, pp. 267--278, 2009.

\bibitem{blackmore2010minimum}
L.~Blackmore, B.~A{\c{c}}ikme{\c{s}}e, and D.~P. Scharf,
  ``Minimum-landing-error powered-descent guidance for {M}ars landing using
  convex optimization,'' \emph{Journal of Guidance, Control, and Dynamics},
  vol.~33, no.~4, pp. 1161--1171, 2010.

\bibitem{hogan2016feedback}
F.~R. Hogan and A.~Rodriguez, ``Feedback control of the pusher-slider system:
  {A} story of hybrid and underactuated contact dynamics,'' in
  \emph{Algorithmic Foundations of Robotics}.\hskip 1em plus 0.5em minus
  0.4em\relax Springer, 2020, pp. 800--815.

\bibitem{lecun2006tutorial}
Y.~LeCun, S.~Chopra, R.~Hadsell, M.~Ranzato, and F.~Jie~Huang, ``A tutorial on
  energy-based learning,'' \emph{Predicting Structured Data}, vol.~1, no.~0,
  2006.

\bibitem{schlick2010molecular}
T.~Schlick, \emph{Molecular Modeling and Simulation: An Interdisciplinary
  Guide}.\hskip 1em plus 0.5em minus 0.4em\relax Springer, 2010, vol.~2.

\end{thebibliography}
	
\end{document}